\documentclass{article}
\usepackage[preprint]{neurips_2024}
\usepackage[T1]{fontenc}
\usepackage{lmodern}
\usepackage{microtype}
\usepackage{amsmath, amssymb}
\usepackage{graphicx}
\usepackage{float}
\usepackage{enumitem}
\usepackage{comment}
\usepackage[hidelinks]{hyperref}
\usepackage{tikz}
\usetikzlibrary{matrix, calc, positioning, arrows.meta}

\usepackage{listings}
\usepackage{xcolor}

\definecolor{codegray}{rgb}{0.5,0.5,0.5}
\definecolor{codegreen}{rgb}{0,0.6,0}
\definecolor{codepurple}{rgb}{0.58,0,0.82}
\definecolor{backcolour}{rgb}{0.95,0.95,0.92}

\definecolor{colabKeyword}{rgb}{1,0.6,0.6}
\definecolor{colabString}{rgb}{0.19,0.55,0.91}
\definecolor{colabComment}{rgb}{0.5,0.5,0.5}
\definecolor{colabBackground}{rgb}{0.17,0.17,0.17}
\definecolor{colabText}{rgb}{0.77,0.77,0.77}

\lstdefinestyle{colabstyle}{
    backgroundcolor=\color{colabBackground},
    commentstyle=\color{colabComment}\ttfamily,
    keywordstyle=\color{colabKeyword}\ttfamily,
    stringstyle=\color{colabString}\ttfamily,
    basicstyle=\color{colabText}\ttfamily\footnotesize,
    numbers=left,
    numberstyle=\tiny\color{colabText},
    numbersep=5pt,
    breaklines=true,
    breakatwhitespace=false,
    keepspaces=true,
    showspaces=false,
    showstringspaces=false,
    showtabs=false,
    tabsize=2,
    captionpos=b,
    language=Python
}
\title{Predictive Coding Implemented as a Hardware Substrate Computes Gradients faster than Backpropagation}
\author{
Timothy Oh \\
}
\date{November 2026}

\begin{document}
\maketitle

\begin{abstract}
Backpropagation has enabled modern deep learning but is difficult to realize as an online, 
fully distributed hardware system due to global error propagation, separate phases, 
and centralized memory. Predictive coding is an alternative learning algorithm with similar 
dynamics to backpropagation that arises from local prediction error dynamics. 
This paper shows that predictive coding can be instantiated as a discrete time digital 
hardware substrate, with local memory and computation, no instruction set architecture,
and one physical core for each predictive-coding neuron. Each core stores its own activity and weights,
communicates with the cores that it's connected to, and performs local gradient updates.
By updating neuron cores concurrently, the architecture computes a network update in $O(J+K)$
clock cycles. For dense layers of width \(n\), this is \(O(n)\) cycles per update, compared with \(O(n^2)\) 
connection operations processed serially.
This result is retained with scaling to larger predictive coding networks, demonstrating
a decisive theoretical advantage for predictive coding instantiated in hardware.
\end{abstract}

\section{Introduction}

Modern machine learning systems are typically trained by backpropagation \cite{rumelhart1986learning}. 
Although this has led to enormous success on digital machines, tasks humans perform effortlessly remain challenging for machines \cite{moravec1988mind}. 
This observation is called Moravec's Paradox, and tasks considered difficult by Moravec's paradox include sensorimotor tasks, robotics, commonsense reasoning, and creative leaps. 
The central hypothesis is that Moravec's paradox is caused, in large part, by the von Neumann bottleneck and weight movement between a centralized memory and the compute units. 
GPUs have massively closed this gap, but network parameters and activations still move between memory and compute units during execution, incurring latency and energy costs.

Predictive coding offers an alternative formulation in which there exist two phases: inference and learning \cite{rao1999predictive}. 
In the canonical form of predictive coding, during the inference phase, errors are passed around between the layers in the predictive coding network until the network reaches an equilibrium. 
Then the learning phase happens, which updates the weights. 
Predictive coding can approximate backpropagation, and particular
formulations approach exact backpropagation gradients under specific
assumptions \cite{whittington2017approximation,millidge2022predictive}.
A newer version of predictive coding, called incremental predictive coding \cite{salvatori2022incremental} (iPC), 
formulates a version that combines the inference and learning phase per update, 
losing the exact backpropagation equivalent weight update but still approximating it. 
Even though the iPC update is not equivalent to backpropagation, the predictive coding network can still learn useful behavior.


There are currently debates on whether predictive coding has computational advantages over backpropagation. 
Backpropagation has been extensively optimized on the CPU-GPU stack, whereas predictive coding typically requires simulating an inference phase, adding computation to the learning process.
It might be possible that the locality of predictive coding as well as other biologically plausible features could be exploited to produce computational advantages on neuromorphic hardware. 
Do such advantages exist?

The answer is yes: parallel neuron cores allow the architecture to compute a network update in a number of clock cycles linear in the maximum fan-in and fan-out. This provides a latency advantage over serial processing of network connections on traditional hardware.
The implementation and experiments are available at
\url{https://github.com/alskaf1293/digitalpredictivecoding}.
\section{Predictive Coding}

Predictive coding can be described by its per neuron dynamics. For neuron $i$ in layer $l$ in the predictive coding network, the neuron stores its own state $x^{l}_{i}$, prediction $\mu^{l}_{i}$, error $\epsilon^{l}_{i}$. The dynamics evolve as follows
\cite{salvatori2022incremental}

\begin{align}
    \mu_i^\ell
        &= \sum_j \theta_{ij}^{\ell} f(x_j^{\ell+1}) + b_i^\ell \\
    \epsilon_i^\ell
        &= x_i^\ell - \mu_i^\ell \\
    \Delta x_i^\ell
        &= \gamma \left(
            -\epsilon_i^\ell
            + f'(x_i^\ell)
              \sum_k \theta_{ki}^{\ell-1}\epsilon_k^{\ell-1}
        \right) \\
    \Delta\theta_{ij}^{\ell}
        &= \alpha\epsilon_i^\ell f(x_j^{\ell+1})
\end{align}

\section{A digital predictive coding network}
The predictive coding hardware is implemented as a network via smaller, locally interacting predictive coding cores. The implementation of the predictive coding core involves a finite state machine to implement equations 1 through 4. Using the locality property of predictive coding, we can represent every neuron in predictive coding as a physical neuron core. The efficiency of predictive coding can be realized by making a few design choices, some of which differ from conventional hardware in many ways:

\vspace{0.5em}
\begin{center}
\begin{tikzpicture}[
  scale=0.9, transform shape,
  sq/.style={draw=black, thick, minimum size=1.25cm},
  sqcore/.style={draw=black, line width=2.5pt, minimum size=1.25cm},
  edge/.style={draw=black, line width=0.35pt},
  edgecore/.style={draw=black, line width=2.5pt},
  lab/.style={font=\small},
  thetalab/.style={font=\small, fill=white, inner sep=1.5pt}
]

\def\dy{2.1}
\def\dx{3.0}
\coordinate (C) at (0,0);

\node[sqcore] (L1) at ($(C)+(\dx,  \dy)$) {};
\node[sqcore] (L2) at ($(C)+(\dx,   0)$) {};
\node[sqcore] (L3) at ($(C)+(\dx, -\dy)$) {};

\node[sq]     (M1) at ($(C)+(0, 1.5*\dy)$) {};
\node[sqcore] (M2) at ($(C)+(0, 0.5*\dy)$) {};
\node[sq]     (M3) at ($(C)+(0,-0.5*\dy)$) {};
\node[sq]     (M4) at ($(C)+(0,-1.5*\dy)$) {};

\node[sqcore] (R1) at ($(C)+(-\dx,  2*\dy)$) {};
\node[sqcore] (R2) at ($(C)+(-\dx,  1*\dy)$) {};
\node[sqcore] (R3) at ($(C)+(-\dx,   0)$) {};
\node[sqcore] (R4) at ($(C)+(-\dx, -1*\dy)$) {};
\node[sqcore] (R5) at ($(C)+(-\dx, -2*\dy)$) {};

\node[lab, right=2mm of L1] {$x^{(\ell+1)}_{1}$};
\node[lab, right=2mm of L2] {$x^{(\ell+1)}_{2}$};
\node[lab, right=2mm of L3] {$x^{(\ell+1)}_{3}$};

\node[lab] at (M2.center) {$x^{(\ell)}_{i}$};

\node[lab, left=2mm of R1] {$x^{(\ell-1)}_{1}$};
\node[lab, left=2mm of R2] {$x^{(\ell-1)}_{2}$};
\node[lab, left=2mm of R3] {$x^{(\ell-1)}_{3}$};
\node[lab, left=2mm of R4] {$x^{(\ell-1)}_{4}$};
\node[lab, left=2mm of R5] {$x^{(\ell-1)}_{5}$};

\foreach \i in {1,2,3} {
  \foreach \j in {1,2,3,4} {
    \draw[edge] (L\i.west) -- (M\j.east);
  }
}
\foreach \i in {1,2,3,4} {
  \foreach \j in {1,2,3,4,5} {
    \draw[edge] (M\i.west) -- (R\j.east);
  }
}

\foreach \j in {1,2,3} {
  \draw[edgecore] (L\j.west) -- (M2.east);
}
\foreach \k in {1,2,3,4,5} {
  \draw[edgecore] (M2.west) -- (R\k.east);
}

\foreach \j in {1,2,3} {
  \node[thetalab] at ($(L\j.west)!0.55!(M2.east)+(0,4pt)$) {$\theta^{(\ell)}_{i\j}$};
}
\foreach \k in {1,2,3,4,5} {
  \node[thetalab] at ($(M2.west)!0.55!(R\k.east)+(0,4pt)$) {$\theta^{(\ell-1)}_{\k i}$};
}

\node[lab, below=4mm of L3] {$\ell + 1$};
\node[lab, below=4mm of M4] {$\ell$};
\node[lab, below=4mm of R5] {$\ell - 1$};

\end{tikzpicture}
\end{center}
\vspace{0.5em}

\begin{enumerate}
\item A neuron in the predictive coding algorithm represents a physical core.
Every core connects locally to its neighboring cores. 
This architecture utilizes the locality of predictive coding. 
Assume that neuron $i$ in layer $l$ is connected to $J$ neurons in the layer above and $K$ neurons in the layer below. 
The core utilizes $2J$ connections, for receiving the activities $x_{j}$ from the layer above and sending signals back, 
as well as sending activities to the K neurons in the layer below and receiving them. 
In total, there are at least $2J + 2K$ pins that enter/exit each neuron core, 
plus additional ones for clock, reset pin, etc. 
The specific architecture details are prone to change, as adjustments are made to optimize the design further.

\item All cores operate in parallel. 
If a tick means that all the stages of the FSM complete in $X$ cycles, 
then for every $X$ cycles, the activities of unclamped cores and, when learning is enabled, the weights of the cores are updated with a step in the $\Delta x$ and $\Delta \theta$ gradient direction. 
Because memory and compute are co-located, 
there is no weight movement like in von Neumann architectures. 
This is where the efficiency comes from.

\item Mortality. 
A key feature of von Neumann machines is that hardware and software are separate, 
which allows instructions that are specified on one machine to be replicated on another machine. 
The software is separate from the hardware. Hinton proposes mortal computation as an alternative in which learning
adapts to the unique physical properties of an individual hardware
system, so its learned parameters are tied to that particular system
\cite{hinton2022forward}. This is a crucial property that makes general-purpose digital machines different from the human brain, and arguably advantageous. 
We use mortality to emphasize the removal of the hardware-software boundary: the predictive coding update equations are implemented directly in the hardware, rather than specified as software instructions running on a general-purpose processor.
In contrast to CPUs and GPUs, a core in this hardware only does one thing. 
While this might seem like it'd render the hardware useless due to its lack of flexibility and programmability, 
it should be emphasized that the efficiencies come about by making this tradeoff. 
The fact that the functionality is restricted to the predictive coding updates allows a core to be much more compact in its machinery. 
This provides also a concrete advantage over immortal systems: the activities and weights of a neuron core (except at the boundaries of a predictive coding network) do not have to be read out in order for the system to work. 
This allows a single PC neuron to be miniaturized further by reducing pin congestion and would allow them to be scaled further by saving area on silicon.

\end{enumerate}

\begin{figure}[ht]
    \centering
    \includegraphics[width=\linewidth]{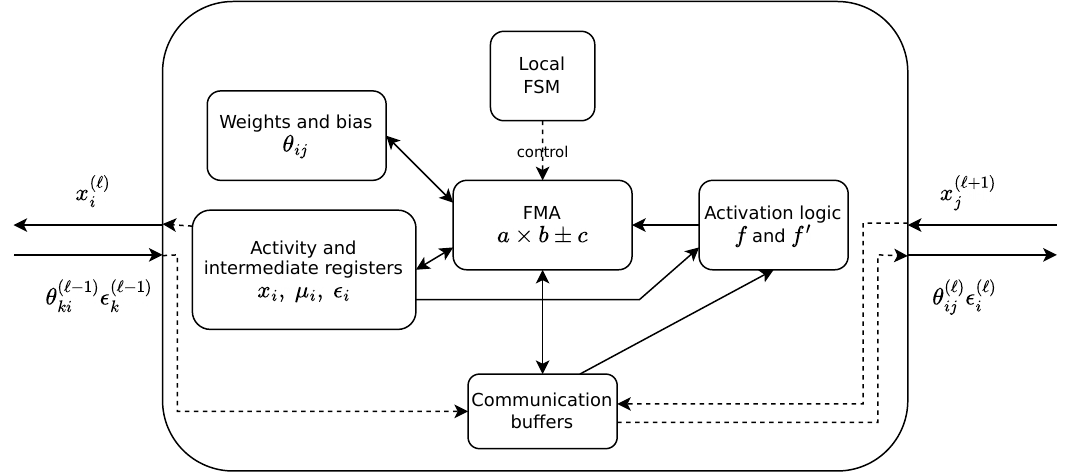}
    \caption{High-level microarchitecture of a predictive coding neuron core.}
    \label{fig:neuron_core}
\end{figure}

\begin{enumerate}
\setcounter{enumi}{3}
\item Online learning. 
This implementation uses iPC, which differs from the canonical implementation of predictive coding by merging the inference and the learning phase 
such that both can be executed in a loop, removing the need for a central controller that explicitly switches the network between inference and learning phases.

\item Every core owns its own finite state machine. The completion of a single cycle of all the FSM stages refers to a tick. A tick involves computing the $\Delta x$ and $\Delta \theta$ for that neuron, which involves the inference and learning phases of predictive coding

\item Every neuron owns its own FMA unit, which it reuses to compute
predictions, feedback contributions, and updates.

\end{enumerate}


Because there are $J$ weights that every neuron core stores, the processing per tick for the weight update step is $O(J)$. \\

The neuron core also receives $K$ back inputs that need to be summed to compute the state update. More specifically, within every tick, each neuron in the layer below computes a contribution

\begin{equation}
    b_{ki} = \theta^{(l-1)}_{ki} \epsilon^{(l-1)}_{k}
\end{equation}

The neuron receives $K$ of these values and sums them, which takes $K$ cycles since the summation is done through one FMA. \\

These are the two operations that dominate the FSM stages. 
The tick latency for a single core is $O(J+K)$.
Since the cores operate in parallel, the network tick latency is
$O(J_{\max}+K_{\max})$, where the maxima are taken across all cores.
For dense layers of width $n$, this corresponds to $O(n)$ cycles per
tick, compared with $O(n^2)$ connection operations processed serially.
It is an interesting question whether the brain implements something similar.

\section{Convergence Results}

We evaluate learning on five regression functions generated by teacher
networks, ranging from a linear mapping to multilayer nonlinear functions.
A predictive coding network is trained separately on each function using
a floating-point Python model of the update equations. Each experiment uses
the same set of randomly sampled inputs for training and evaluation.

\begin{figure}[ht]
    \centering
    \includegraphics[width=0.85\linewidth]{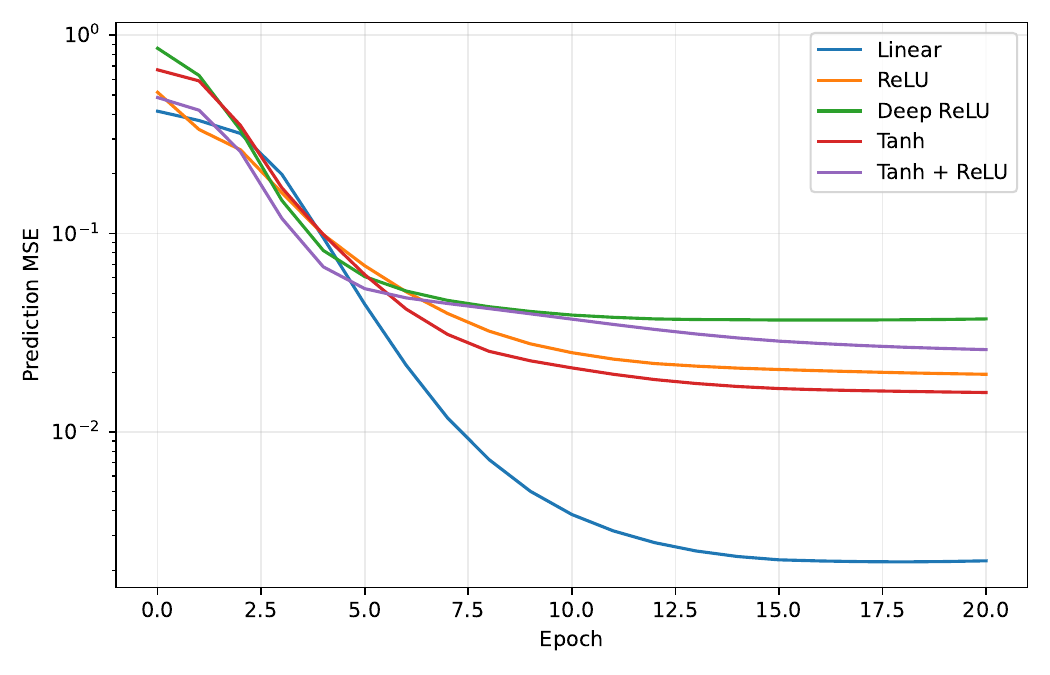}
    \caption{Prediction error on randomly sampled training points for five teacher functions.}
    \label{fig:convergence}
\end{figure}

Figure~\ref{fig:convergence} shows that prediction error decreases on all
five functions. These results demonstrate learning on the tested training
points using the Python model.

\section{RTL to GDS}

The neuron core and a small network composed of these cores were taken
through a physical design flow to produce GDS layouts. The network layout
passed the reported layout-versus-schematic and design-rule checks. 

\section{Moravec's Paradox}

There are many good reasons to suspect that this hardware would be better suited for tasks considered difficult by Moravec's Paradox. 
There is evidence that predictive coding is better suited for online, low data, batch-1 learning \cite{song2024inferring}. Furthermore, due to property 4, the phases between inference and learning no longer exist, meaning that the full utilization of the predictive coding hardware will come from a system that is continually learning and requires frequent weight updates. \\

In a conventional robotics system that utilizes backpropagation, an agent is trained on a training distribution

$$
\left( s', s,a\right) \sim P_{\text{train}}
$$

and during deployment, the agent encounters ice, gravel, or uneven terrain, and the observations come from a different distribution

$$
P_{\text{test}}\left(s'\mid s,a \right) \neq P_{\text{train}}\left( s'\mid s,a\right)
$$


There have been several attempts to get around out of distribution (OOD) problem in the RL literature. 

Domain randomization accepts the static inference deployment phase by exposing the training distribution to as many scenarios as possible, but still incurs the problem when leaving that distribution \cite{peng2018simtoreal,leibovich2022validate}. 
Classical adaptation is effective and computationally efficient for estimating a small number of parameters (eg. friction, mass, external disturbances) but faces limitations when facing dynamics not adequately represented by the controller's predefined model \cite{rohrs1985robustness}. 
A natural solution to this would be to represent the controller via neural networks, as continual RL does \cite{khetarpal2022continual}.
However, this still encounters the OOD problem again due to the computational demands of the von Neumann stack.


The transition model was used as an example. 
However, the value, actor, and critic would all benefit from online learning. 
The actor $\pi_\phi(a \mid s)$ selects actions; a policy trained on dry ground might continue selecting actions that cause a slip on ice. The critic $Q_\psi(s,a)$ estimates the expected return of actions; under new dynamics previously favorable actions might become dangerous or ineffective. 
The OOD problem can affect other pretrained neural networks in a reinforcement learning system. By supporting local, online updates, this hardware could provide a substrate for adapting these networks as the environment changes.
\section{Discussion}
Mead's seminal paper \cite{mead1990neuromorphic} on neuromorphics outlines a few key ideas that depart from the digital computing paradigm. 
One of these ideas is local and parallel compute, which this paper's architecture implements along with in memory compute being another motivation of neuromorphic computing.
The rest of the ideas that have not been implemented suggest future improvements. These include bringing digital predictive coding neurons into the analog domain to reduce neuron size, complexity, and increase efficiency in exchange for noise and precision.
Information could be represented with analog signals like voltage or current instead of digital bits.
The per-neuron predictive coding equations could more naturally be implemented as an analog circuit using Kirchhoff's current law for summation, voltage-conductance multiplication for weighted inputs, 
and capacitive integration for continuous-time activity dynamics. \\

Also, the digital results should theoretically hold for larger networks, although this would need to be empirically established.
These efforts, combined with a neuromorphic ecosystem, would eventually enable the computational advantages of predictive coding to be realized. 
These improvements could also be the precursor to a computing system that is fully local, parallel, and continuously learning as it interacts with its environment.

\bibliographystyle{plain}
\bibliography{references}

\end{document}